\documentclass[11pt]{article}
\usepackage{pixta}
\usepackage{ragged2e}
\usepackage{graphicx}
\graphicspath{ {./figures/} }
\usepackage{amsmath,amssymb}
\usepackage{bbold}
\usepackage{booktabs}
\usepackage{threeparttable}
\usepackage{float}
\usepackage[style=ieee, url=false, isbn=false, doi=false, eprint=false]{biblatex}
\usepackage{hyperref} 
\hypersetup{
    colorlinks=true,
    linkcolor=blue,
    filecolor=magenta,
    urlcolor=cyan,
}
\usepackage{tabularx}
\usepackage{mathtools}
\DeclarePairedDelimiter\floor{\lfloor}{\rfloor}

\usepackage[inline]{enumitem}

\title{MAGNeto: An Efficient Deep Learning Method\par for the Extractive Tags Summarization Problem}

\author{
  Hieu Trong Phung$^{1,2}$ \\
  {\tt hieu.phungtrong@pixta.co.jp} \\
  \And
  Anh Tuan Vu$^{1}$ \\
  {\tt tuananh.vu@pixta.co.jp} \\
  \AND
  Tung Dinh Nguyen$^{1}$ \\
  {\tt tung.nguyendinh@pixta.co.jp} \\
  \And
  Lam Thanh Do$^{1,2}$ \\
  {\tt lam.dothanh@pixta.co.jp} \\
  \And
  Giang Nam Ngo$^{1}$ \\
  {\tt giang.ngonam@pixta.co.jp} \\
  \AND
  Trung Thanh Tran$^{1}$ \\
  {\tt trung.tranthanh@pixta.co.jp} \\
  \And
  Ngoc C. L\^{e}$^{1,2}$ \\
  {\tt lechingoc@yahoo.com} \\
  \AND
  \\
  \normalsize{$^{1}$PIXTA Vietnam, 8th Floor, Truong Thinh Building, Phung Chi Kien, Cau Giay District, Hanoi, Vietnam.}
  \\
  \normalsize{$^{2}$Hanoi University of Science and Technology, 1 Dai Co Viet Road, Ha Noi, Viet Nam.}
}

\begin{document}
\sloppy 
\maketitle

\begin{abstract}
\textit{
In this work, we study a new image annotation task named Extractive Tags Summarization (ETS).
The goal is to extract important tags from the context lying in an image and its corresponding tags.
We adjust some state-of-the-art deep learning models to utilize both visual and textual information.
Our proposed solution consists of different widely used blocks like convolutional and self-attention layers, together with a novel idea of combining auxiliary loss functions and the gating mechanism to glue and elevate these fundamental components and form a unified architecture.
Besides, we introduce a loss function that aims to reduce the imbalance of the training data and a simple but effective data augmentation technique dedicated to alleviates the effect of outliers on the final results.
Last but not least, we explore an unsupervised pre-training strategy to further boost the performance of the model by making use of the abundant amount of available unlabeled data.
Our model shows the good results as 90\% $F_\text{1}$ score on the public NUS-WIDE benchmark, and 50\% $F_\text{1}$ score on a noisy large-scale real-world private dataset.
Source code for reproducing the experiments is publicly available at: \url{https://github.com/pixta-dev/labteam}
}
\end{abstract}

\textbf{Keywords:} Image Annotation, Extractive Tags Summarization, Deep Neural Network, CNN, Self-Attention

\section{Introduction}

\begin{figure}[ht]
\includegraphics[width=\linewidth]{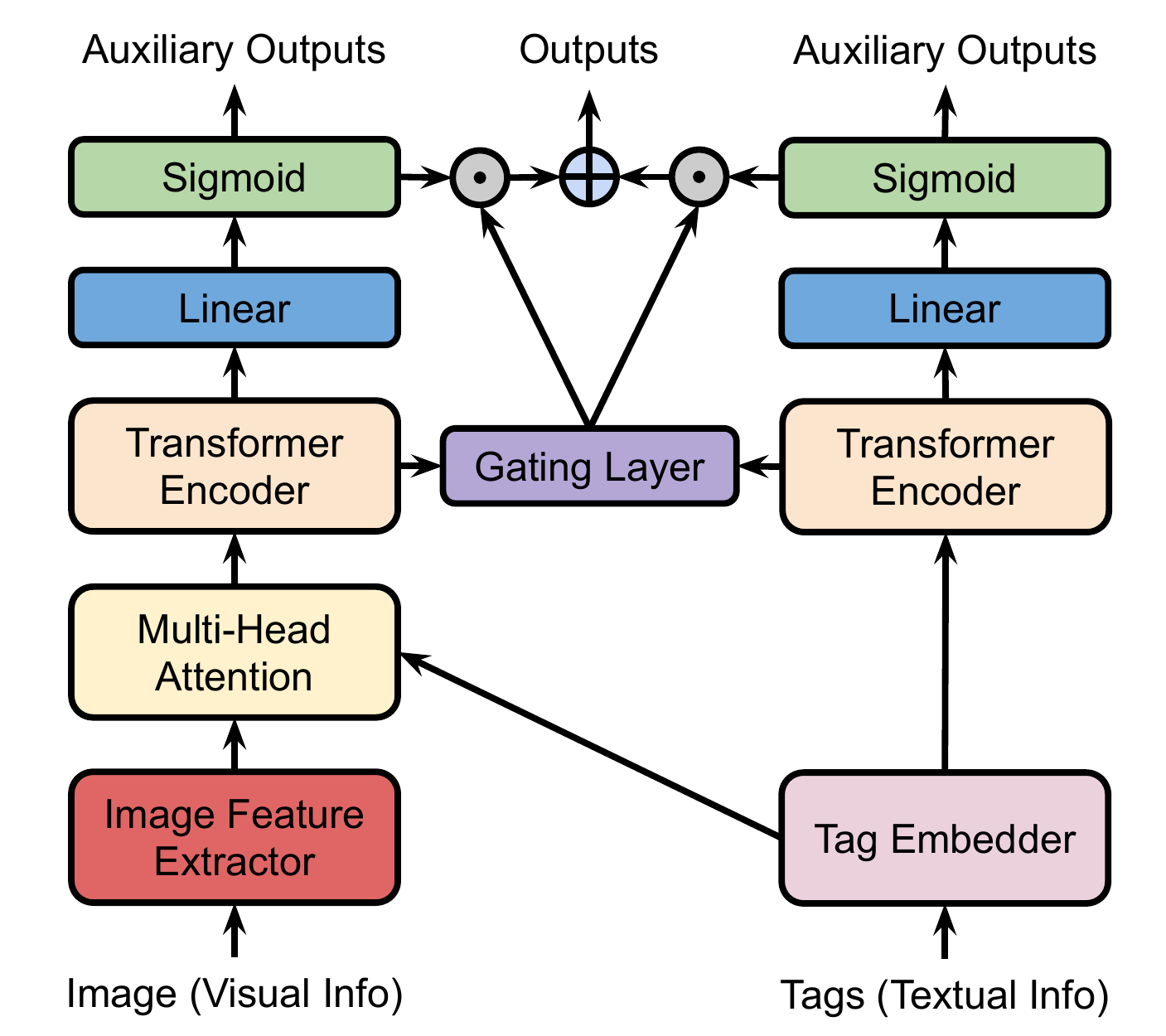}
\caption{A two-stream network utilizing both visual (left) and textual (right) information to solve ETS task.}
\label{fig:magneto_architecture}
\end{figure}

The goal of ETS task is to shorten the list of tags corresponding to a digital image while keeping the representativity.
The process provides not only an effective way to quickly understand the image's content but also possibly serve other downstream tasks such as image retrieval~\cite{Wu2013}, caption generation~\cite{Chen2015,You2016,Johnson2016}, object detection~\cite{Hariharan2014}, etc.
Undoubtedly, deep learning is a powerful tool for numerous problems these days with the backing of abundant data as well as the advances in hardware technology. In this research, we compose some deep learning models to combine computer vision (CV) and natural language processing (NLP) tasks to employ both visual and textual information.

Convolutional neural network (CNN) \cite{LeCun1995} has shown to be a suitable architecture for extracting visual features from an image. In fact, it can be applied to most CV tasks as long as having enough training data and computational power.

To the NLP side, in \cite{Vaswani2017}, the authors introduced the Transformer architecture, an attention-based model with the help of self-attention mechanism, has beaten recurrent models like \cite{Hochreiter1997,Chung2014} to become a very effective solution for NLP tasks.

In this work, we proposed a unified architecture that employs both convolutional layers and the Transformer encoder with the help of auxiliary loss function \cite{Szegedy2015,Zhao2017} and a gating mechanism to solve the ETS task in an efficient way to employ visual features from the image as well as textual features from tags (Figure \ref{fig:magneto_architecture}). The following is the description of the ETS framework.

Assuming that each image has already been assigned all the relevant tags required to adequately describe the content of this image, our job is to ``summarize" the original set of tags to a more compact form which still being able to help understand the main point presented, \textit{i.e.}, only are important tags kept. Along with that, we propose a simple but effective deep learning method to deal with ETS task that can extensively exploit the information in both image and tag.

In Section \ref{secMethods}, we describe the model architecture as well as the used loss function. Section \ref{secDataAndExperimentResults} is devoted to the data and conducted experiments. Some details about the setting of experiments are left in Section \ref{secExperimentSettings}. In Section \ref{secConclusion}, we discuss the results as well as some potential considerations in the future. But first, in Section \ref{secRelatedWork}, we describe some researches related to the issue.

\section{Related Work}
\label{secRelatedWork}
Image annotation is the task of labeling images with tags to reflect the images' visual contents. Due to its significance, many pieces of research have been carried in various manners.

\textbf{Tags as the target.}
When the labels are the related tags to the images, a greedy search-based tag order determination approach \cite{Qian2014} could be used to re-tag images with the tag relevance and diversity problem as well. In \cite{Wu2017}, the authors modified the conditional determinantal point process (DPP) algorithm as DPP sampling with weighted semantic paths dealing with a new image annotation task, named diverse image annotation (DIA), retrieving a limited number of tags needed to cover as much as useful information about the input image as possible. Besides the diversity, a recent publication \cite{Verma2019} cares about the missing label problem and proposed a new $k$-nearest neighbor (k-NN) based algorithm for the issue.
With creativity, even modern technology like GANs \cite{Goodfellow2014} can be utilized to handle the image annotation problems \cite{Wu2018, Ke2019}.

\textbf{Deep learning in computer vision.}
Since the publication of LeNet-5 \cite{LeCun1998} in the 1990’s, convolutional neural networks have gone through a long journey to become an effective, or rather the best, general architecture for automatically extracting useful features from images.
Due to the excessive computation cost and the disinterests of researchers in this kind of architecture, there was not any significant research on CNN in many years after LeNet-5.
It is not until AlexNet \cite{Krizhevsky2012} that popularized convolutional networks in the computer vision field by beating the second runner-up of the ImageNet ILSVRC 2012 challenge \cite{Russakovsky2015} with a huge margin.
Year after year, new model architectures have been proposed to surpass the existed solutions.
We have ZFNet \cite{Zeiler2014}, GoogLeNet \cite{Szegedy2015}, NiN \cite{Lin2013}, VGGNet \cite{Simonyan2014}, ResNet \cite{He2016}, and many more \cite{Iandola2016,Szegedy2016a,Szegedy2016,Hu2018,Zhang2018,Huang2017,Xie2017,Chollet2017,Howard2017,Ma2018,Sandler2018,Howard2019,Han2020,Zhang2020}, continuing the loop of development.

\textbf{Deep learning in natural language processing.}
Comes to the NLP field, deep learning also has a great influence on its development.
It has been a long time since recurrent-based models, such as long short-term memory (LSTM) \cite{Hochreiter1997}, and gated recurrent unit (GRU) \cite{Chung2014}, dominated existed solutions for various NLP tasks like language modeling and machine translation \cite{Sutskever2014,Bahdanau2014,Cho2014}.
With the newborn of attention-based models like Transformer \cite{Vaswani2017}, the boundary of deep learning in natural language processing field has been pushed further.
Training massive deep learning language models \cite{Radford2018,Devlin2018,Radford2019,Zellers2019,Shoeybi2019,Rosset2019} has become the current trend in the field with the mark of a 175 billion
parameter language model named GPT-3 \cite{Brown2020} recently.

\section{Methods}
\label{secMethods}
In this section, we propose the models used in this research.
\subsection{Model Architecture}
\begin{figure*}[ht]
\includegraphics[width=\linewidth]{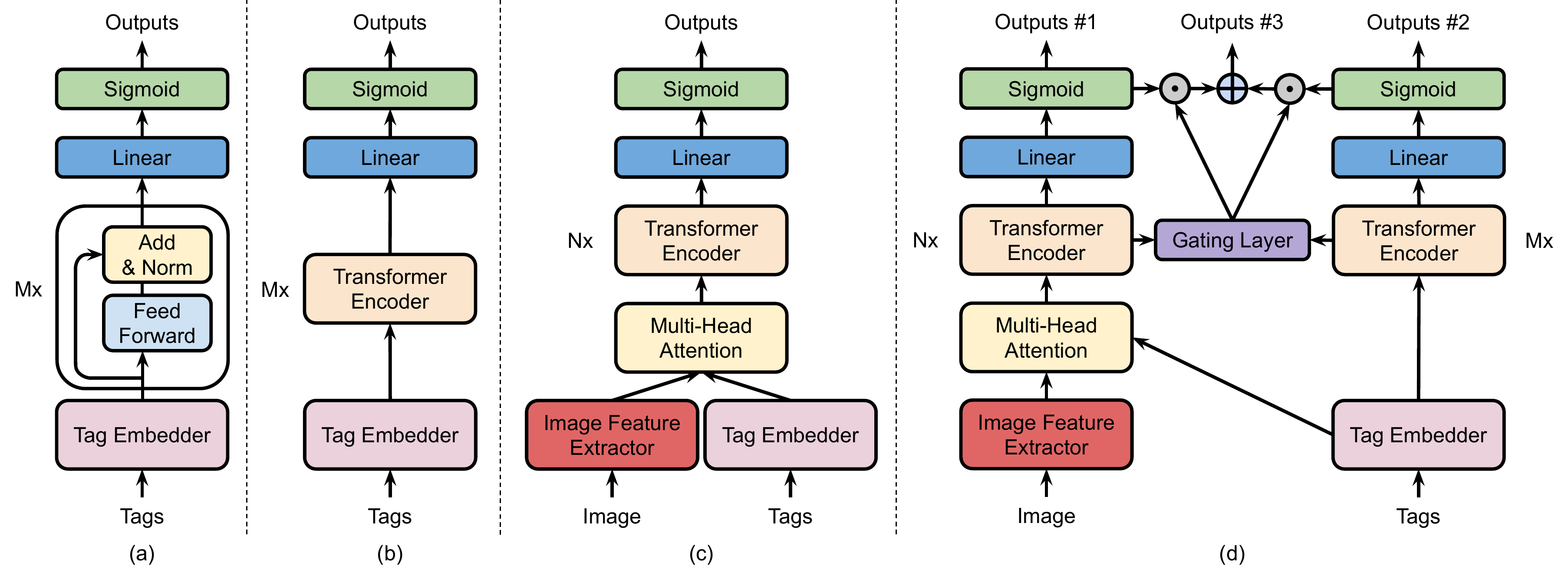}
\caption{From the left to the right: (a) The baseline model that mainly composed of the feed-forward network and only requires tags for the input, (b) the baseline model that utilizes the Transformer encoder to handle tags' features, (c) the baseline model that utilizes the Transformer encoder to handle tags' and image's features, (d) MAGNeto architecture.}
\label{fig:model_architectures}
\end{figure*}

Our proposed solution, named MAGNeto (Multi-Auxiliary with Gating Network), is composed of two streams:
\begin{enumerate*}[label=(\arabic*)]
    \item Tag-stream which only uses tag's information to solve ETS task, and
    \item image-stream which requires both image's and tag's information.
\end{enumerate*}
Above all is a gating mechanism taking the role of fusing the outputs of two streams (Figure \ref{fig:model_architectures}).

\textbf{Image feature extractor.}
The input image's features are extracted in this step and any CNN architecture can be used for this purpose. In this research, we modify ResNet \cite{He2016} as the following.
First, we drop the top global average pooling \cite{Lin2013} and fully connected layers.
Second, a $1\times1$ convolutional layer is plugged on top of the network to get the desired shape of the output feature maps, followed by a batch-norm \cite{Ioffe2015} layer.
Finally, an $(N, N, d_{model})$ tensor that correspond to $N\times N$ grid regions with $d_{model}$ features is added.

\textbf{Tag embedder.}
The tags are vectorized here. PyTorch framework provides us the trainable embedding module, or layer for convenience, to learn the suitable embedding models. Any pre-trained models such as Word2Vec \cite{Mikolov2013a,Mikolov2013} or GloVe \cite{Pennington2014} can be used here.

\textbf{Transformer encoder.}
One of the most important parts of the whole architecture, the Transformer encoder, composed of identical layers built with multi-head self-attention mechanism and position-wise fully connected feed-forward network.
This component figures out the relations among feature vectors, \textit{e.g.}, the vectors that corresponding to tags or grid regions in the image's feature maps.
As the MAGNeto architecture described in Figure \ref{fig:model_architectures}, there are two blocks of the Transformer encoder: (1) stacked of $M$ identical layers for the tag-stream in the right and (2) stacked of $N$ identical layers for the image-stream in the left.

\textbf{Multi-head attention layer.}
In this subsection, we describe the method combining visual features from the image and textual features from tags. Since the target of the ETS task is to determine which tags should be kept, the tags' feature vectors should be projected on the image feature vector space ($\mathbb{R}^{d_{model}}\to\mathbb{R}^{d_{model}}$) by utilizing the attention (Scaled dot-product attention \cite{Vaswani2017}) mechanism.
The tags' feature vectors are \textit{queries} while image regions' vectors play as \textit{keys} and \textit{values}, \textit{i.e.}, each tag vector ``looks" at the whole grid regions of the image feature maps and the duty of the attention mechanism is to guide the tag to which regions should be attended.
The desired outputs would be the fusion of the image's and tags' information, which then fed into the Transformer encoder to extract more abstract features.

\textbf{Gating mechanism.}
The gating layer is a special part of our model merging the outputs of the tag-stream and the image-stream to form the final result.
The sub-layers of this special block are listed in Table \ref{tab:gating_layer}.

\begin{table}[ht]
    \centering
    \begin{threeparttable}
        \begin{tabular}{@{}lll@{}}
        \toprule
        Layer   & \multicolumn{1}{c}{In shape} & \multicolumn{1}{c}{Out shape} \\ \midrule
        Dropout & $(bs, l, 2\times d_{model})$ & $(bs, l, 2\times d_{model})$  \\
        FC      & $(bs, l, 2\times d_{model})$ & $(bs, l, d_{ff})$             \\
        ReLU    & $(bs, l, d_{ff})$            & $(bs, l, d_{ff})$             \\
        Dropout & $(bs, l, d_{ff})$            & $(bs, l, d_{ff})$             \\
        FC      & $(bs, l, d_{ff})$            & $(bs, l, 1)$                  \\
        Squeeze & $(bs, l, 1)$                 & $(bs, l)$                     \\
        Sigmoid & $(bs, l)$                    & $(bs, l)$                     \\ \bottomrule
        \end{tabular}
        \begin{tablenotes}
            \footnotesize
            \item \textit{$bs$: Batch-size.}
            \item \textit{$l$: The maximum number of tags per item. To serve the batching purpose, all the items must have a fix number of tags, here is $l$ tags.  Empty slots are filled up with zero values, i.e., we use padding for items that do not have long enough sets of tags.}
            \item \textit{$d_{model}$: The number of expected features in the encoder inputs.}
            \item \textit{$d_{ff}$: The dimension of the feed-forward network.}
        \end{tablenotes}
    \end{threeparttable}
    \caption{Gating layer's units. (The flow of the data is from top to the bottom.)}
    \label{tab:gating_layer}
\end{table}

Let $\mathcal{O}_{it} = [o_{it}^0, o_{it}^1, \dots, o_{it}^{l-1}]^\mathsf{T},$ $o_{it}^{i}\in[0, 1]$ be the output of the image-stream and $\mathcal{O}_{t} = [o_{t}^0, o_{t}^1, \dots, o_{t}^{l-1}]^\mathsf{T},$ $o_{t}^{i}\in[0, 1]$ be of the tag-stream.
The output of each stream can play standalone as an answer to the problem. However, fusing these outputs to get the ultimate decision usually results in a better outcome.
The gating layer uses the intermediate feature vectors, the outputs of two Transformer encoders concatenated along the last axis, which results in $(2 \times d_{model})$-dimensional feature vectors, in the both streams of the network to return a vector $\mathcal{A} = [\alpha_{0}, \alpha_{1}, \dots, \alpha_{l-1}]^\mathsf{T},$ $\alpha_{i}\in[0, 1]$ for each item.
The final output of the model is calculated as following.

\begin{equation}
\begin{aligned}
  \mathcal{O}_\text{Final} = \mathcal{A}^\mathsf{T} \odot \mathcal{O}_{it} + (\mathbb{1} - \mathcal{A}^\mathsf{T}) \odot \mathcal{O}_{t},
\end{aligned}
\label{eq:final_output}
\end{equation}

where $\odot$ is the element-wise (or point-wise) multiplication.

\subsection{Loss Function}
\textbf{Overcoming class imbalance problem.}
Binary Cross-Entropy (BCE) loss can be seen as the default option for any binary classification problems.

\begin{equation}
\begin{aligned}
  \mathcal{L}_\text{BCE}(y_{true}, y_{pred}) =
  & - y_{true} \log(y_{pred}) \\
  & - (1 - y_{true})\log(1 - y_{pred}),
\end{aligned}
\label{eq:bce_loss}
\end{equation}

where $y_{true}$ is the binary indicator (0 or 1) of the class label and $y_{pred}$ is the predicted probability.

In literature, when dealing with the class imbalance between positive and negative examples, Dice loss seems to be a much better option dedicated to handling this problem.

\begin{equation}
  \mathcal{L}_\text{Dice}(y_{true}, y_{pred}) = 1 - \frac{2|y_{true} \cap y_{pred}|}{|y_{true}| + |y_{pred}|}.
\label{eq:dice_loss}
\end{equation}

This special loss function and its variations have been widely used in various image segmentation problems for both 2D and 3D inputs \cite{Milletari2016,Drozdzal2016,Taghanaki2019,Isensee2018,Scandalea2019,Fidon2018,Sudre2017,Zhu2019}.

Since the ETS task can be assumed as a 1D image segmentation problem, Dice loss's properties perfectly fit our purpose, preventing class imbalance.
The Dice loss \eqref{eq:dice_loss} is combined with the traditional BCE loss \eqref{eq:bce_loss} to form the BCE-Dice objective function as the following.

\begin{equation}
  \mathcal{L}_\text{BCE-Dice} = \mathcal{L}_\text{BCE} + \mathcal{L}_\text{Dice}.
\label{eq:bce_dice_loss}
\end{equation}

\begin{figure}[ht]
\includegraphics[width=\linewidth]{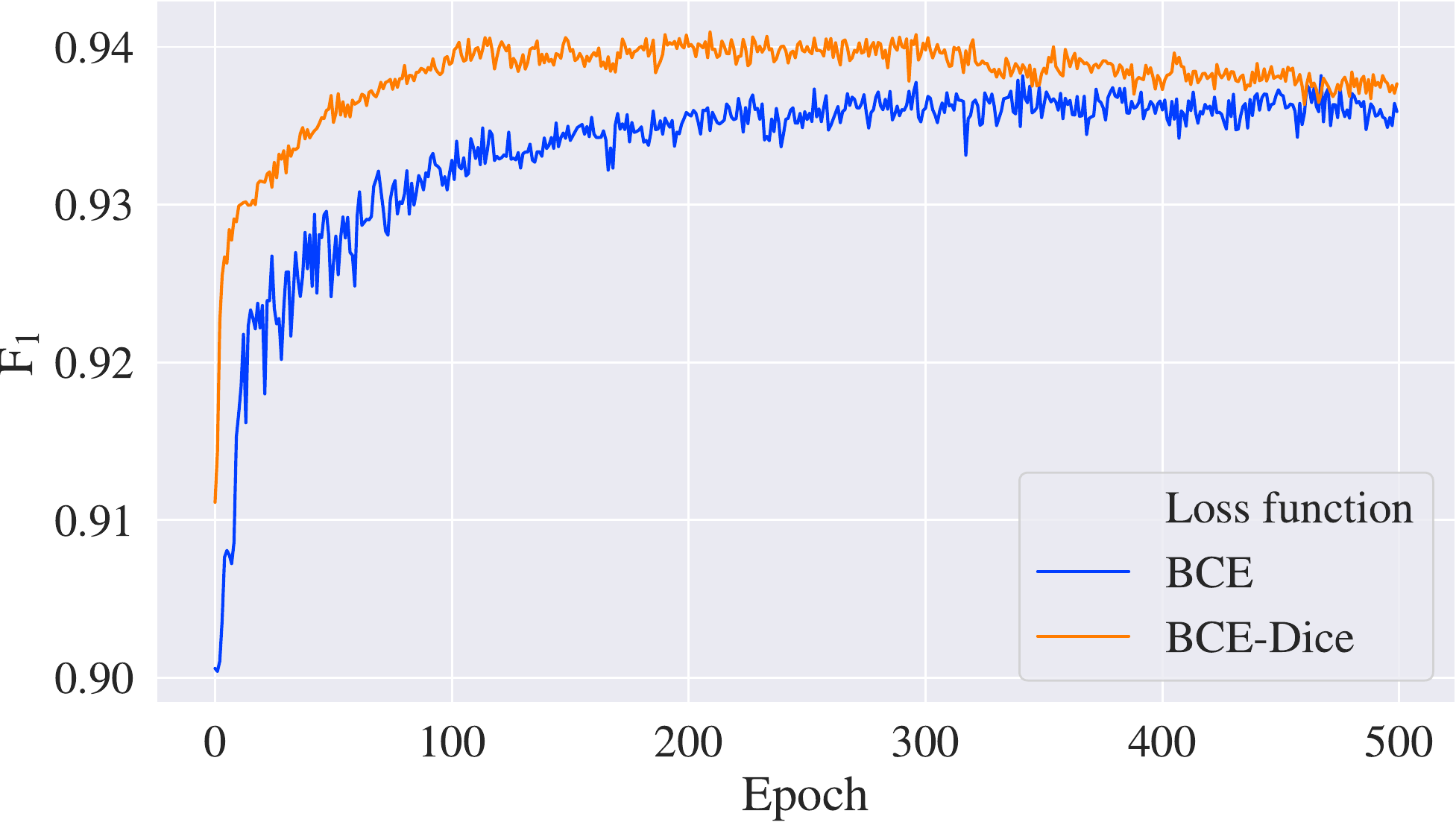}
\caption{BCE loss vs. BCE-Dice loss [NUS-WIDE].}
\label{fig:nus_wide_loss_comparison_bce_vs_bcedice}
\end{figure}

In our experiments, the BCE-Dice loss helps not only to speed up the convergence rate but also to increase the performance of the model. (Figure~\ref{fig:nus_wide_loss_comparison_bce_vs_bcedice}).

\begin{figure*}[ht]
\includegraphics[width=\linewidth]{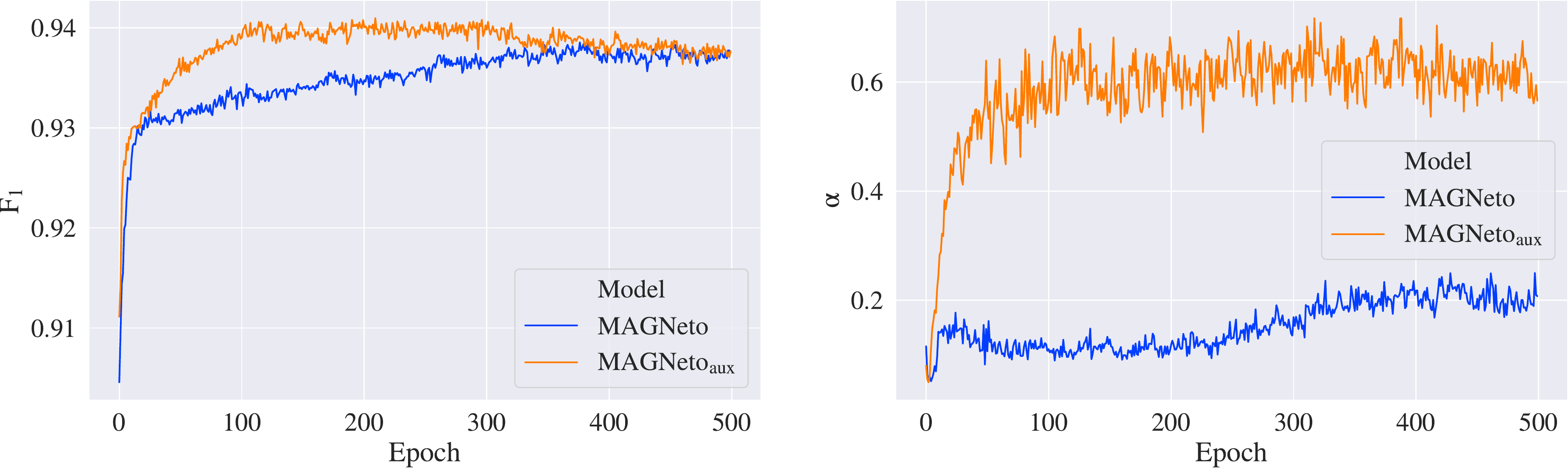}
\caption{Without and with auxiliary loss functions [NUS-WIDE].}
\label{fig:nus_wide_aux_loss}
\end{figure*}

\textbf{Auxiliary loss functions.}
In general, since the tags often contain less information than the image, the tag stream tends to converge much faster than the image stream and it would be difficult for the model to converge. Hence, besides the main loss function guides the training process of the whole model, two auxiliary loss functions for the image stream and tag stream have been added to the model. 
When one stream learns faster than the other, the $\alpha$ values tend to be skewed.
In some extreme cases, the slower-converging stream can be ignored altogether.
Figure \ref{fig:nus_wide_aux_loss} depicts the learning process of the two versions, with and without auxiliary loss functions.

For simplification, the final objective of MAGNeto are computed as:
\begin{equation}
\begin{aligned}
  \mathcal{L}_\text{Total}
  & = \mathcal{L}_{g} + \mathcal{L}^1_{aux} + \mathcal{L}^2_{aux} \\
  & = \mathcal{L}_{g} + \mathcal{L}_{it} + \mathcal{L}_{t},
\end{aligned}
\label{eq:loss_total}
\end{equation}
where $\mathcal{L}_{g}$ is the loss function for the final output after going through the gating mechanism, two auxiliary loss functions $\mathcal{L}^1_{aux}$ ($\mathcal{L}_{it}$) and $\mathcal{L}^2_{aux}$ ($\mathcal{L}_{t}$) are employed to guide the training processes of the image-stream and tag-stream, correspondingly.

\section{Data and Experiment Results}
\label{secDataAndExperimentResults}

In this section, we introduce two sets of training data and show some experiments related to the model architecture. Some details of the experiment settings are left in the next section.

\subsection{Training Data}
\textbf{NUS-WIDE dataset.}
To serve the target of ETS task, we perform some pre-processing steps to the public NUS-WIDE dataset \cite{ChuaJuly8-102009} as the following.
\begin{enumerate}
    \item For each item, \textit{i.e.}, image, only keeping the tags included in the 81 concepts that can be used for evaluation provided by the National University of Singapore.
    \item Filtering out all the items that cannot satisfy the condition: Having the set of tags that cannot cover all the annotated concepts in the ground-truth.
    \item Filtering out all the items that do not have any tag left after applying the two above pre-processing steps.
\end{enumerate}

\begin{figure*}[ht]
\includegraphics[width=\linewidth]{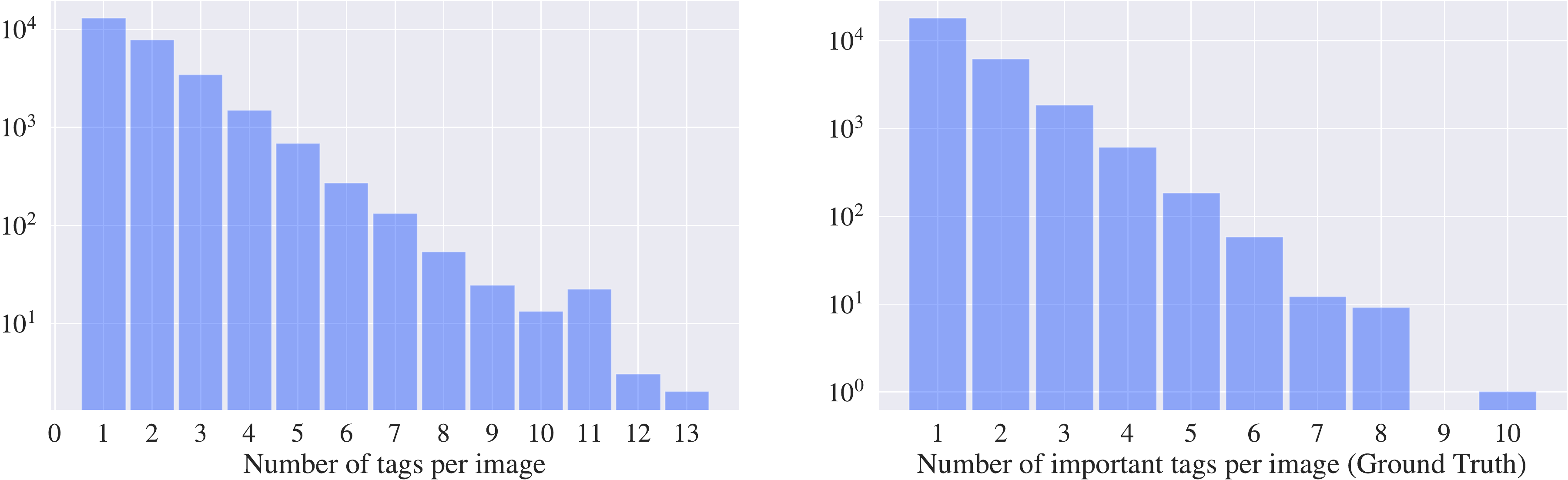}
\caption{Data distribution [NUS-WIDE].}
\label{fig:nus_wide_dist}
\end{figure*}

From 269,648 images, after going through all the pre-processing steps, we have obtained 26,559 satisfied items for the ETS task.
The distribution of the new dataset is shown in Figure \ref{fig:nus_wide_dist}.

\textit{Note: All the experiments in this section are demonstrated by using the NUS-WIDE public dataset.}

\textbf{Large-scale dataset.}
To test the capacity of the proposed solution in the next section, we have generated a large-scale real-world dataset, abbreviate as LS for convenience, which composed of about 700,000 fully annotated images with 24,056 concepts.

\subsection{Baseline Models}
Before coming up with the multi-auxiliary with gating idea, we have implemented several architectures from the simplest composed of fully-connected layers to big networks using attention mechanism (Figure~\ref{fig:model_architectures}).

\textbf{Feed-forward networks with residual connections.}
This is the most straightforward architecture using only tag information to determine the output. We name this architecture FF for later references.
Examining this architecture shows us that the relations among tags are extremely significant to find the important concepts for each image.

\textbf{The Transformer encoder for tag features.}
This architecture nearly the same as the FF mentioned above, except changing the feed-forward networks to the Transformer encoder.
We name this architecture TF-t for later references.

\textbf{The Transformer encoder for image with tag features.}
Until now, we only use the tag information to do ETS task, but how about the image, the good source of information to decide which tags should be kept?
To utilize the image information, we combine it with tag features by projecting the tags' feature vectors to the image feature vector space with the attention mechanism.
Then, we use the Transformer encoder to learn the relations among these intermediate features to get more meaningful ones before classifying the input tags.
We name this architecture TF-it for later references.

\subsection{Proposed Solution}
\begin{figure}[ht]
\includegraphics[width=\linewidth]{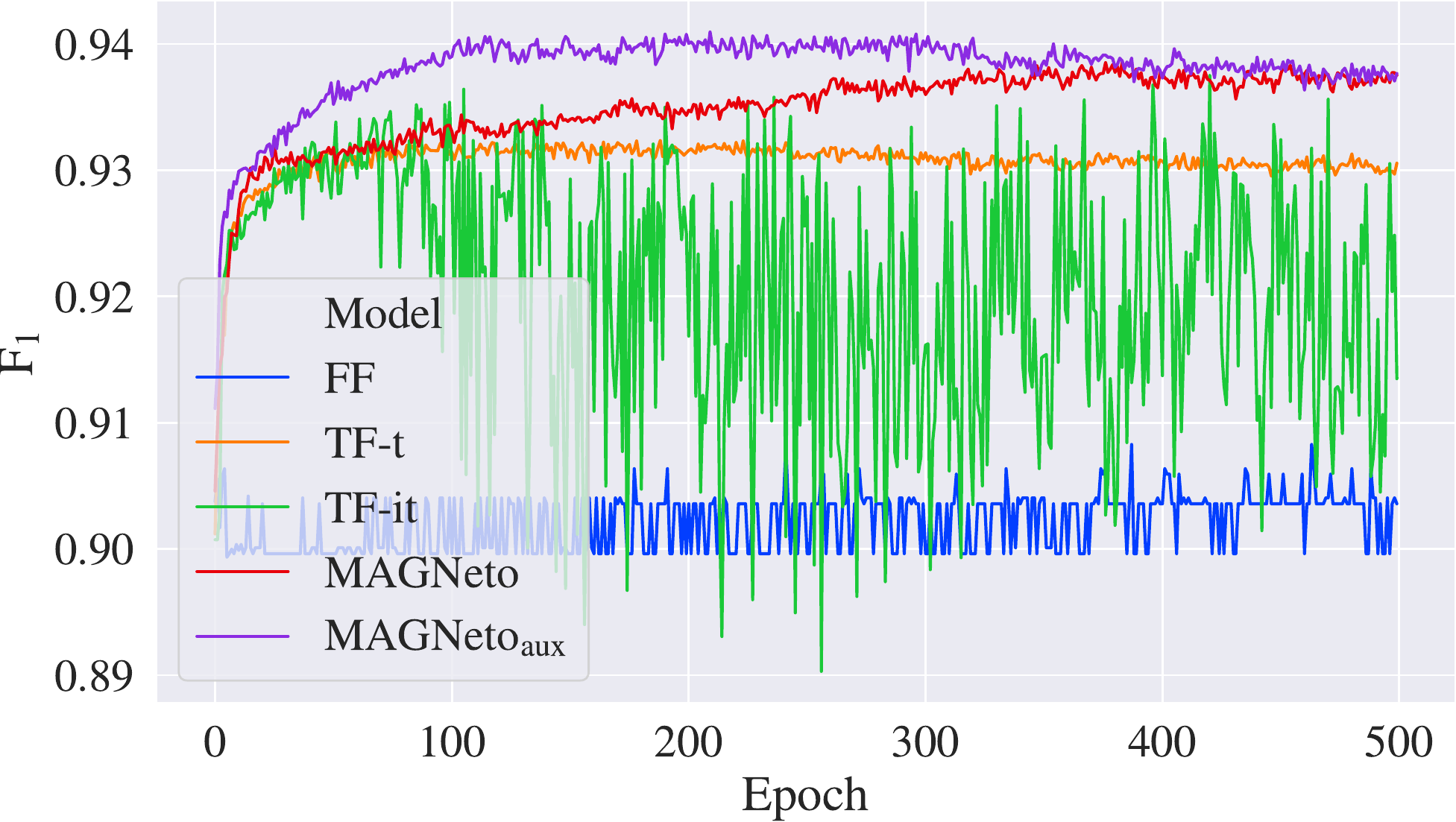}
\caption{$\mathrm{F_1}$ comparison [NUS-WIDE].}
\label{fig:nus_wide_model_comparison_f1}
\end{figure}

Our proposed architecture has two streams and is utilized with the gating mechanism.
Despite the importance of image information, using tag information alone can boost the performance of the model significantly.
Obviously, having two experts in hands seems much better, one entirely relies on tag features and one makes use of all available resources.
After having had the outputs of both streams, the gating layer comes into play, fusing the outputs of these two experts, image-stream and tag-stream, to form the final decision on which tags should be marked as important.
Based on the $\mathrm{F_1}$ score, MAGNeto beats all the baseline models. The results are described in Figure~\ref{fig:nus_wide_model_comparison_f1} and Table~\ref{tab:metrics}.

\begin{table}[ht]
    \centering
    \begin{tabular}{l c c c} \toprule
        Model & Prec (\%) & Recall (\%) & $\mathrm{F_1}$ (\%) \\ \midrule
        FF  & 90.5 & 97.1 & 90.8 \\
        TF-t  & 93.4 & 98.1 & 93.2 \\
        TF-it & 94.4 & \textbf{100.0} & 93.8 \\ \midrule
        MAGNeto & 93.9 & 97.6 & 93.9 \\
        $\mathrm{MAGNeto_{aux}}$ & \textbf{94.5} & 96.7 & \textbf{94.1} \\ \bottomrule
    \end{tabular}
    \caption{Evaluation metrics of different architectures.}
    \label{tab:metrics}
\end{table}

\section{Experiment Settings}
\label{secExperimentSettings}

In this section, we describe data augmentation, unsupervised pre-training strategy, and testing model's capacity with LS dataset.

\subsection{Outliers and Tag Data Augmentation}
Although the powerful self-attention mechanism in the Transformer encoder tremendously helps the MAGNeto architecture to extract useful features from the textual information provided by the input tags, it cannot defend the model from the outliers, \textit{i.e.}, inappropriate tags.
In some extreme cases, these unrelated, to the image content, tags could be classified as the important ones.
This effect might be the result of the differences between outliers' vectors and the rest.
To address this issue without performing any pre-filtering methods to the raw input tags, we propose an augmentation technique for the textual content which can be efficiently performed during the training phase.
We name it ``tag adding \& dropping", or TAD for short, augmentation.
Not only does this technique help overcome the outlier problem but also improves the generalization of the model by reducing overfitting.
(As the name suggests, TAD augmentation includes two subprocedures: Tag-adding and tag-dropping.)

\textbf{Tag-adding.}
This is the procedure of randomly picking unique tags from the vocabulary and employing these to extend the original list of tags corresponding to each item.
We have the $\beta$ coefficient responsible for the maximum ratio between the number of new adding tags and of the original tags that labeled as not important.
For example, with $\beta = 0.3$, if an item has a set of 50 tags in which five of them are marked as important tags and the rest, 45 tags, are unimportant, the maximum number of tags we can add up is $A = \floor*{0.3 \times 45} = 13$.
After having the $\beta$ upper bound, we uniformly sample an integer between $[0, A]$ and use it as the true number of tags should be randomly~\footnote{Uniformly sampling.} selected from the vocabulary~\footnote{All original tags must be excluded from the vocabulary prior to executing the tag sampling process.}.

\textbf{Tag-dropping.}
In contrast to the tag-adding procedure, tag-dropping randomly picking unimportant tags from the original set to remove them from the list.
We also have the $\hat{\beta}$ coefficient to limit the number of tags being dropped.
In practice, we tend to use $\hat{\beta}$ equal to $\beta$; however, these are two independent hyper-parameters, thus, different values could be used.

To check the effect of outliers on the model, we first randomly add inappropriate tags to the validation set and evaluate the model; then, we count the number of outliers picked as important tags in the final outputs.
The results are described in Table~\ref{tab:performance_comparison}.

\begin{table}[ht]
    \centering
    \begin{threeparttable}
        \begin{tabular}{@{}ccccc@{}}
        \toprule
        $\beta$ & $\hat{\beta}$ & Pre-trained\tnote{\textdagger} & Outlier\tnote{\ddag} (\%) & F1 (\%) \\ \midrule
        0.0     & 0.0           & No          &       39.34       &    93.24     \\
        0.5     & 0.5           & No          &       28.60       &    93.25     \\
        0.0     & 0.0           & Yes         &       24.98       &    94.27     \\
        0.5     & 0.5           & Yes         &       \textbf{20.53}       &    \textbf{94.30}     \\ \bottomrule
        \end{tabular}
        \begin{tablenotes}
            \footnotesize
            \item[\textdagger] Whether or not using unsupervised pre-training weights.
            \item[\ddag] The percentage of items affected by outliers.
        \end{tablenotes}
    \end{threeparttable}
    \caption{Performance comparison among configurations [NUS-WIDE].}
    \label{tab:performance_comparison}
\end{table}

Please be aware that when working with NUS-WIDE dataset, the results may not accurately reflect the true performance of the model in reality.
The reason is because of
\begin{enumerate*}[label=(\arabic*)]
    \item the small number of concepts (only 81 concepts),
    \item the small number of tags per item after using the pre-processing and filtering strategies (mainly 1 tag and 13 tags at most),
    \item and the small number of images (26,559 items).
\end{enumerate*}
However, in LS dataset, when we have
\begin{enumerate*}[label=(\arabic*)]
    \item thousands of concepts (24,056 concepts), 
    \item each item have about 43 tags on average,
    \item and having much more labeled training data (about 700,000 fully annotated images),
\end{enumerate*}
the effect of outliers (about 8-10\% of validation items affected by the presentation of the outliers and 0.5-0.8\% after using TAD augmentation) is not severe as in NUS-WIDE dataset.

\subsection{Unsupervised Pre-training Strategy}
These days, unsupervised pre-training has been widely used in different tasks~\cite{Zhang2017,Yu2010,He2013,Ramachandran2016} because of the abundance of unlabeled data when compared with the nicely labeled one.
As a special case of semi-supervised learning, unsupervised pre-training aims at finding a good initialization point while retaining the supervised learning objective.
When being utilized correctly to support the target supervised learning task, unlabeled data can surely leverage the performance of the model~\cite{Erhan2010}.

\begin{figure}[ht]
\includegraphics[width=\linewidth]{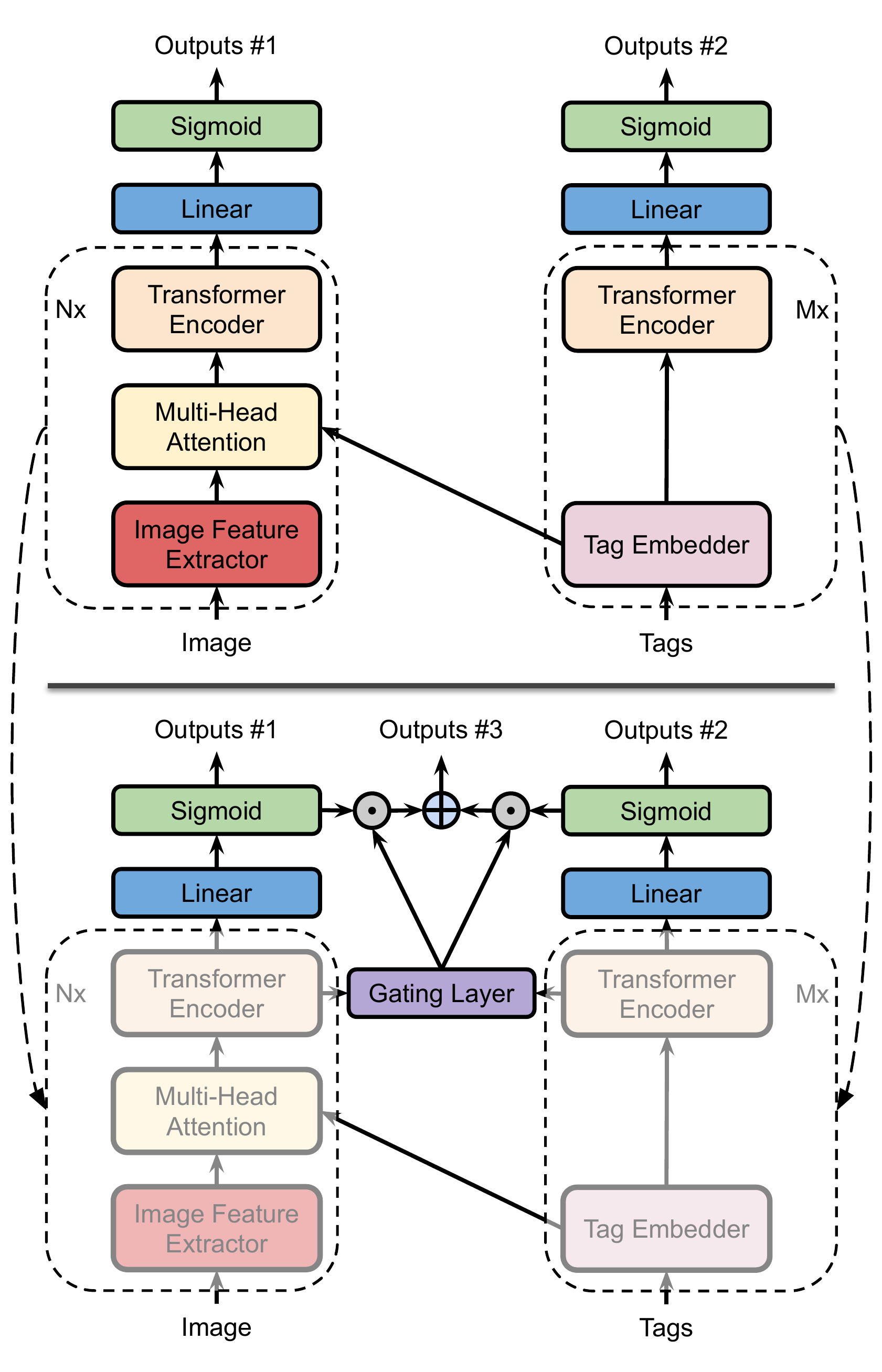}
\caption{The training pipeline composed of unsupervised pre-training (top) and supervised fine-tuning (bottom).}
\label{fig:unsupervised_pretraining_strategy}
\end{figure}

To further improve the performance of the MAGNeto model and make use of unlabeled data available, we propose a simple two-stage training strategy (Figure~\ref{fig:unsupervised_pretraining_strategy}) which is a combination of unsupervised pre-training and supervised fine-tuning.

\textbf{Unsupervised pre-training.}
The model architecture used in this first stage is slightly modified to better fit the purpose of finding a good initialization point for the target model in the supervised fine-tuning stage.
In detail, we remove the gating mechanism in the original MAGNeto architecture entirely and the model becomes a two-input two-output architecture.
(It means the model still has visual and textual inputs and both outputs are on target to attain the same unsupervised pre-training objective with the guidance of the loss function $\mathcal{L}_{u} = \mathcal{L}_{it} + \mathcal{L}_{t}$.)
The objective must be accomplished is detecting outliers, \textit{i.e.}, solving the outlier detection problem; here, outliers referred to the inappropriate tags.
It is simply achieved by randomly adding irrelevant tags to the unlabeled training data and the model's job is to classify relevant and irrelevant tags for each item in the dataset.
After being fitted by a huge amount of unlabeled data, we are ready for the second stage, fine-tuning the model for the target task -- ETS.

\textbf{Supervised fine-tuning.}
At this stage, we come back to our original MAGNeto architecture which is now initialized with the parameters of the unsupervised pre-trained model in the first stage.
Nevertheless, only some specific building blocks are copied to the target model, which are image feature extractor, tag embedder, multi-head attention layer, and two Transformer encoders, while the rest are initialized and trained from scratch.
Note that we do not freeze the pre-trained parameters yet use labeled data for ETS task to fine-tune them together with the whole model.

\begin{figure*}[ht]
\includegraphics[width=\linewidth]{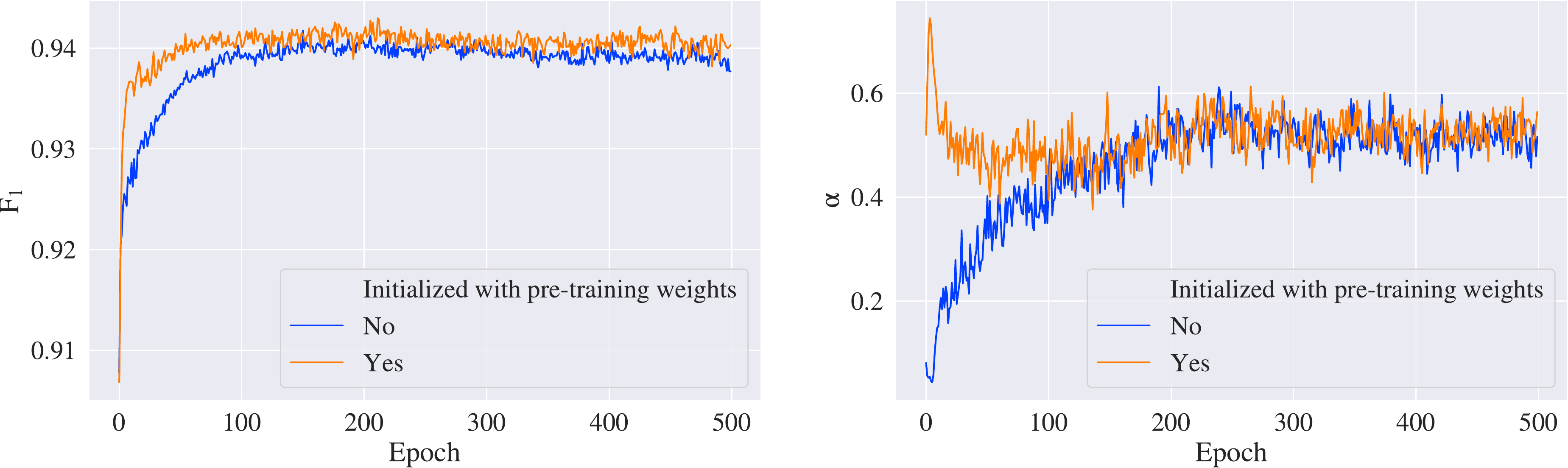}
\caption{Without and with initializing with unsupervised pre-training weights [NUS-WIDE].}
\label{fig:nus_wide_unsupervised_pretraining}
\end{figure*}

As expected, using unsupervised pre-trained weight can stabilize the $\alpha$ values at the beginning of the training process and, more importantly, improve the model's performance (Figure~\ref{fig:nus_wide_unsupervised_pretraining}).
Besides, the unsupervised pre-training strategy alone alleviates the effect of outliers on the target model (Table~\ref{tab:performance_comparison}).
When being coupled with TAD augmentation, the model can be even more robust in ignoring inappropriate tags.

\subsection{Testing Model on a Large-Scale Real-World Dataset}
\begin{figure}[ht]
\includegraphics[width=\linewidth]{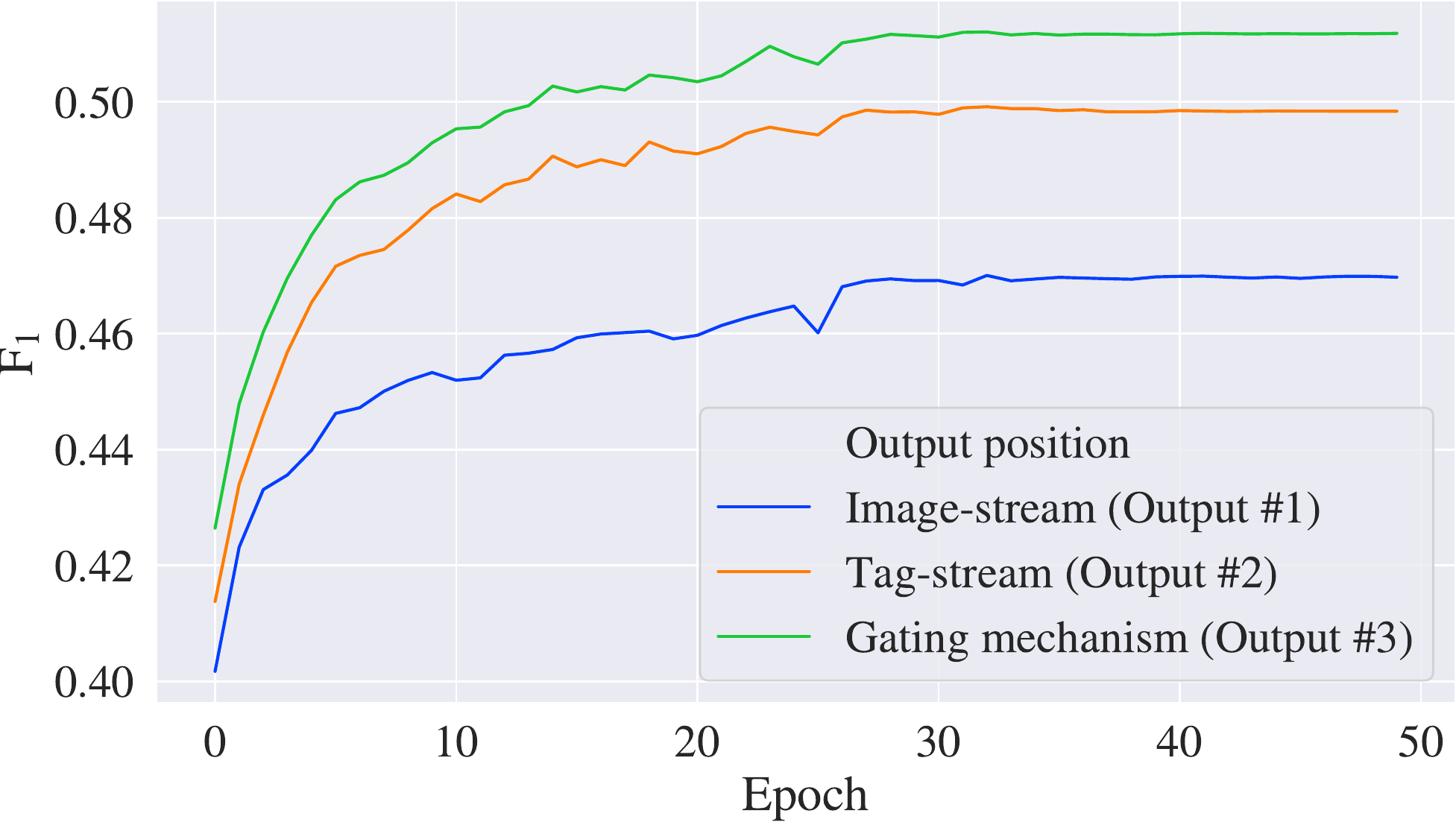}
\caption{$\mathrm{F_1}$ comparison of different output positions [LS].}
\label{fig:large_scale_position_comparison_f1}
\end{figure}

\begin{figure}[ht]
\includegraphics[width=\linewidth]{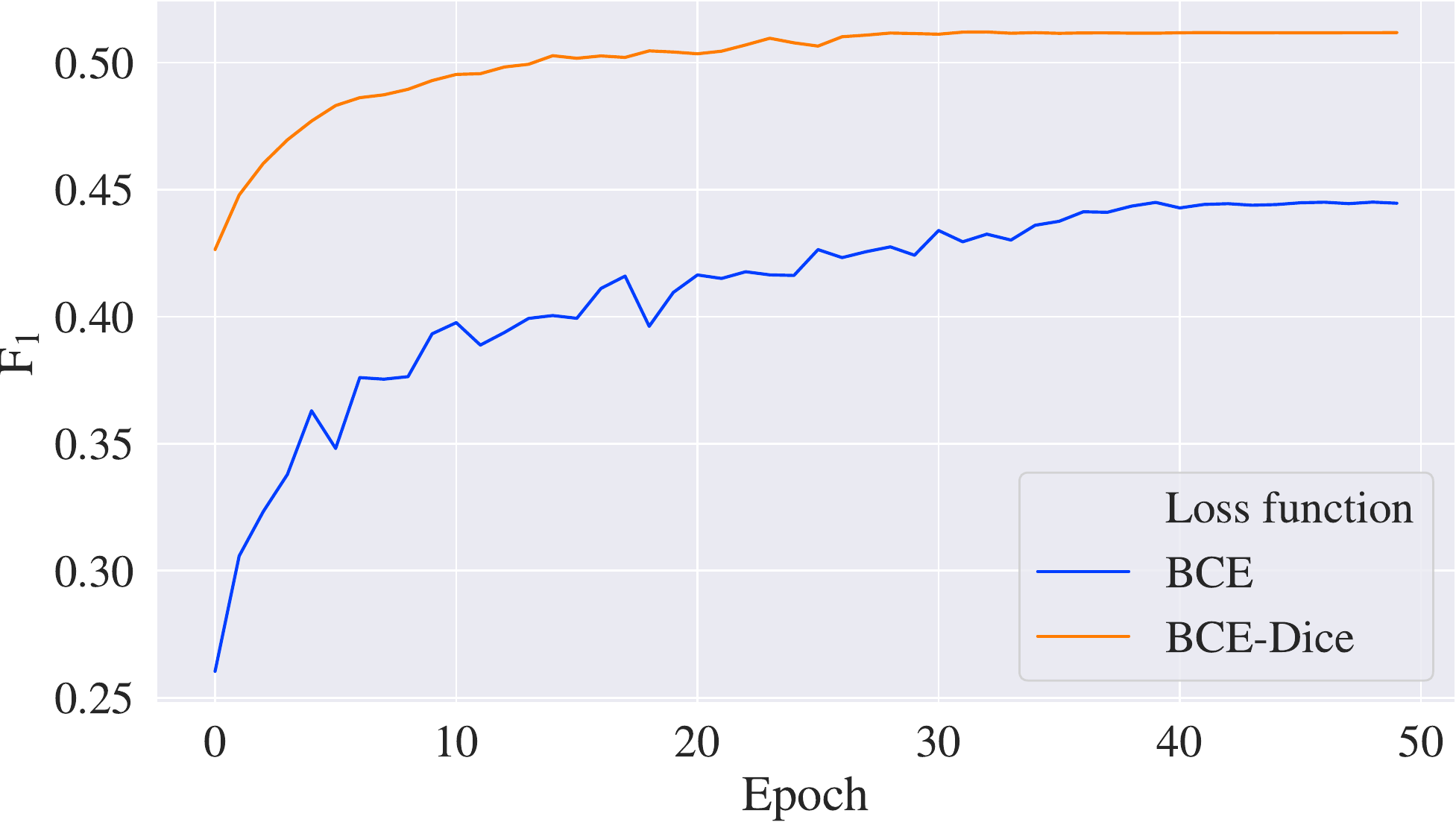}
\caption{$\mathrm{F_1}$ comparison of BCE and BCE-Dice loss [LS].}
\label{fig:large_scale_loss_comparison_bce_vs_bcedice}
\end{figure}

When working with a large-scale real-world dataset, the result is always imperfect when compared with the experiments conducted in the lab.
Additionally, the gaps among experiments are usually much more significant, a few percent of $\mathrm{F_1}$ score, when compared to the results while working with the NUS-WIDE dataset.
However, despite the horrible noise presented in the dataset, the model is still capable of learning the hidden pattern from the data.
Again, the $\mathrm{F_1}$ scores in Figure~\ref{fig:large_scale_position_comparison_f1} and Figure~\ref{fig:large_scale_loss_comparison_bce_vs_bcedice} show the effectiveness of the gating mechanism and the BCE-Dice loss function, respectively.

\section{Conclusion}
\label{secConclusion}
In this work, a new image annotation task called Extractive Tags Summarization (ETS) is studied.
Besides taking advantage of advancements in the deep learning field, we specifically devote our effort to combine these fundamental components together and form a complete model dedicated to solving ETS task. 
After having ablation studies with various baseline models, the final proposed solution named MAGNeto has shown the effectiveness of the gating mechanism when being cooperated with auxiliary loss functions.
Additionally, choosing appropriate objective function has been deeply studied since its great influence on the training process of the whole network; especially, when the imbalance presents in the training dataset.
Moreover, data augmentation techniques and unsupervised pre-training strategies are also taken care of to further boost the performance of the model.

The gating mechanism presented in this work is just our first attempt to fuse the outputs of the two streams.
We have not performed any hyper-parameter turning for this specific layer and even have not decided which are the best input features for it.
Therefore, MAGNeto architecture definitely could be further improved just by leveraging a better gating mechanism.
Another aspect that can improve the model performance is by preventing the imbalance lies in the data.
The BCE-Dice loss introduced in the above section only deals with the imbalance between the number of positive and the negative tags for each input item, \textit{i.e.}, between the number of important tags and the rest.
The balance among the concepts' frequencies is still an open problem.
The last thing that should be paid attention to in future works is the mapping mechanism from the tag feature vector space to the image one, \textit{e.g.}, using a multi-head attention layer as proposed in this work.

\section*{Acknowledgments}
This project is sponsored by PIXTA Inc.
We are grateful to Minh Huu Nguyen, Duong Thuy Do, and Huyen Thi Le for their fruitful comments, corrections, and inspiration. The last author also received the support from Vietnam Institute for Advanced Study in Mathematics in Year 2020.

\printbibliography[heading=bibintoc]

\newpage
\appendix
\section{Training Configurations}
\subsection{Training Data and Batching}
We performed various experiments with our models on 2 datasets: the public NUS-WIDE benchmark and LS -- a large-scale real-world private dataset.
For the demonstrating purpose, we configured \textit{batch-size} $=32$ for NUS-WIDE.
Yet, when dealing with LS, we used \textit{batch-size} $=256$ to make the most of an RTX-2080ti GPU.
Moreover, we use \textit{l} $= 16$ for training with NUS-WIDE dataset, and \textit{l} $= 64$ while working with LS.

\subsection{Image Feature Extractor}
Since the small size of NUS-WIDE dataset after being applied to some pre-processing and filtering steps, ResNet18 was used as the backbone for the image feature extractor.
Additionally, all parameters of the backbone were entirely frozen during the training phase in all experiments.
With a big dataset like LS, we adopted ResNet50 and froze the first 3 residual blocks of the backbone.

\subsection{Transformer Encoder}
Through different empirical studies, it is shown that the Transformer encoder is very easy to fit the data in ETS task; thus, depending on the size of the dataset, the hyper-parameter configurations would be varied.

\textbf{NUS-WIDE.}
We used one block for the encoder of the image-stream ($N = 1$) and two for the tag-stream ($M = 2$). All others hyper-parameters are the same for both streams: \textit{d-model} $= 128$, \textit{heads} $= 4$, \textit{dim-feedforward} $= 512$, \textit{dropout}~\cite{Srivastava2014} $= 0.3$, where \textit{d-model} is the number of expected features in the input, \textit{heads} is the number of heads in the multi-head attention model, \textit{dim-feedforward} is the dimension of the feed-forward networks, and \textit{dropout} is simply the dropout value used for the whole architecture.

\textbf{LS:}
We set $N = 2$, $M = 6$, \textit{d-model} $= 512$, \textit{heads} $= 8$, \textit{dim-feedforward} $= 2048$, \textit{dropout} $= 0.1$.

\subsection{Optimizer}
We used the stochastic gradient descent (SGD) optimizer to train MAGNeto model with $momentum = 0.9$, \textit{learning-rate} $= 10^{-2}$; when dealing with LS, \textit{learning-rate} $= 3\times10^{-2}$ can be used for faster convergence rate.
Furthermore, we also used reduce-lr-on-plateau scheduler with $factor = 0.1$ and $patience = 5$ for the training process with the large-scale dataset.

\subsection{Data Augmentation}
\textbf{Image data augmentation.}
Besides frequently used augmentation techniques for images such as flipping and cropping, we also adopted the ImageNet augmentation policies learned by AutoAugment~\cite{Cubuk2019}.
For image size, we used \textit{input-size} $= 112$ for NUS-WIDE, and \textit{input-size} $= 224$ for LS.

\textbf{Tag data augmentation.}
For TAD augmentation, we used $\beta = 0.5$ and $\hat{\beta} = 0.5$ for NUS-WIDE since the small scale of the dataset.
When training with LS, the two coefficients are smaller with $\beta = 0.3$ and $\hat{\beta} = 0.3$.

\subsection{Hardware}
We trained our model on one machine with a single RTX-2080ti GPU.

\section{Results}
Some results when validating MAGNeto on the NUS-WIDE dataset can be found in Table \ref{tab:nus_wide_results}.

\begin{table*}[ht]
    \centering
    \begin{tabularx}{\textwidth}{p{0.55\linewidth} X X X}
    \toprule
    Image & Tags                                         & Prediction                    & Ground truth                  \\ \midrule
    \raisebox{-0.9\totalheight}{\includegraphics[width=\linewidth]{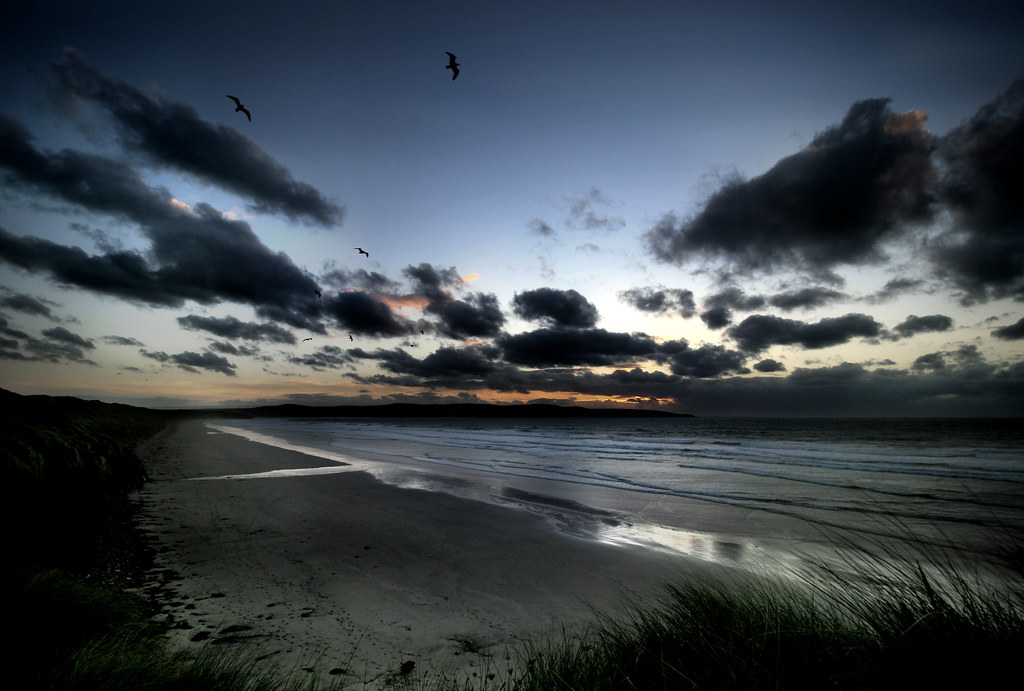}}      & beach\par birds\par clouds\par grass\par ocean\par sky\par sun\par sunset\par water & clouds\par grass\par ocean\par sky\par sunset\par water & beach\par clouds\par ocean\par sky\par water       \\ \midrule
    \raisebox{-0.9\totalheight}{\includegraphics[width=\linewidth]{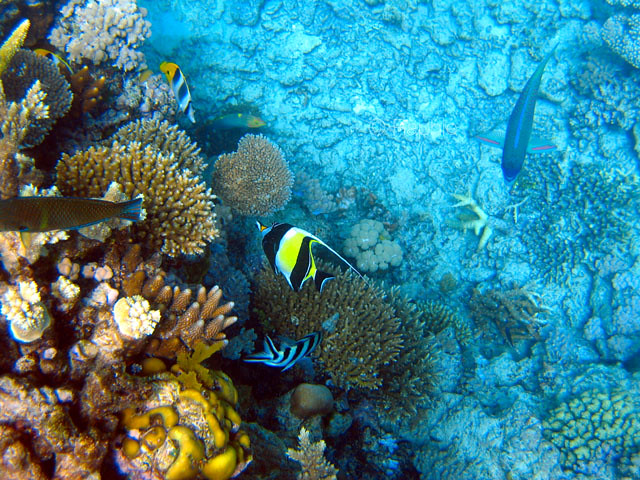}}      & animal\par beach\par coral\par fish\par ocean\par sand\par sun\par water & animal\par coral\par water            & animal\par coral\par fish\par ocean\par water \\ \midrule
    \raisebox{-0.9\totalheight}{\includegraphics[width=\linewidth]{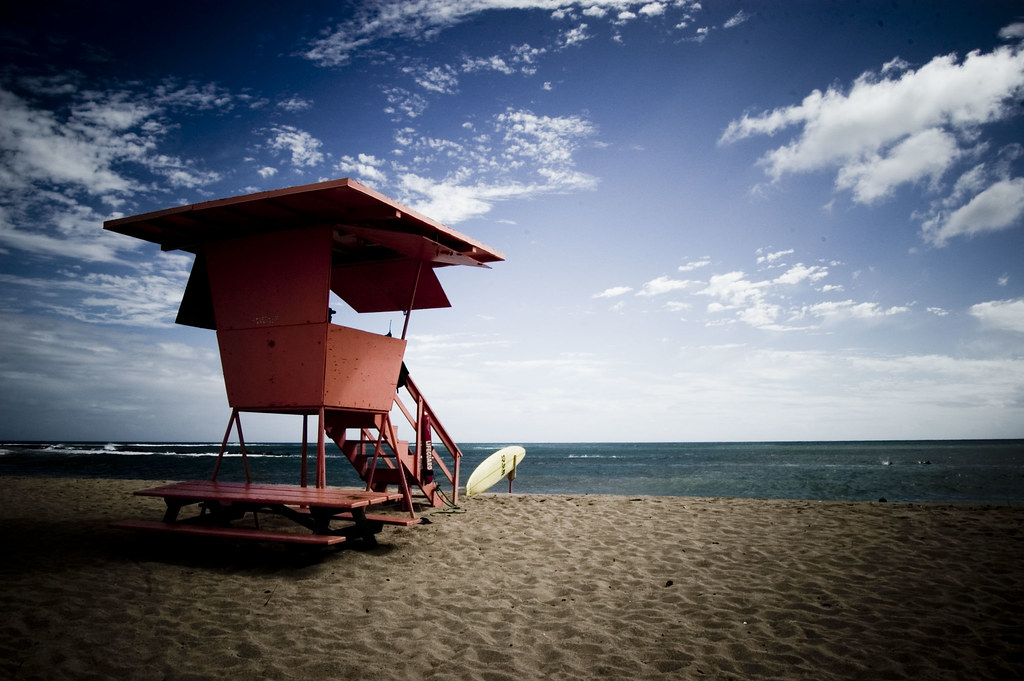}}      & beach\par clouds\par house\par ocean\par sand\par sky\par surf\par water & beach\par clouds\par ocean\par sky\par water            & beach\par clouds\par ocean\par sky\par water \\ \bottomrule
    \end{tabularx}
    \caption{Some results with the NUS-WIDE dataset.}
    \label{tab:nus_wide_results}
\end{table*}

\end{document}